# SYNTHETIC POINT CLOUD GENERATION FOR CLASS SEGMENTATION APPLICATIONS

Maria J. Gonzalez Stefanelli[1], Avi Rajesh Jain[2], Sandeep Kamal Jalui[2] and Dr. Eva Agapaki[1]

[1] M.E. Rinker, Sr. School of Construction Management, University of Florida, USA

[2] H. Wertheim College of Engineering, University of Florida, USA

## Abstract

Maintenance of industrial facilities is a growing hazard due to the cumbersome process needed to identify infrastructure degradation. Digital Twins have the potential to improve maintenance by monitoring the continuous digital representation of infrastructure. However, the time needed to map the existing geometry makes their use prohibitive. We previously developed class segmentation algorithms to automate digital twinning, however a vast amount of annotated point clouds is needed. Currently, synthetic data generation for automated segmentation is non-existent. We used Helios++ to automatically segment point clouds from 3D models. Our research has the potential to pave the ground for efficient industrial class segmentation.

## Introduction

Currently, there is an infrastructure investment gap, and worldwide professional entities are developing solutions to close it. The China Belt and Road is an initiative to evaluate long-term strategies that improve infrastructure in different continents such as Africa, Europe, Australia, Asia, and Latin America (Organization for Economic Co-operation and Development 2018). The strategy is implemented due to the urgent need to alleviate the excessive amounts of global infrastructure failing short by $0.35-0.37 trillion per year (McKinsey 2016). Globally, the improvement of road infrastructure falls into an investment of $0.44 trillion annually. However, the needs will not be met these coming years (Miyamoto & Wu 2018). According to the Department of Energy in the United States, the cost of power outages is up to $70 billion annually leading the country to a required minimum investment of $65 billion to support development of energy infrastructure (House 2021). Infrastructure investments are even more significant for fast growing industries, such as the manufacturing industry (with 35% productivity growth), which creates an outsize economic impact in the United States (Institute 2021). Therefore, the high value of manufacturing assets to our economies necessitates the need to properly maintain the industrial sector.

The United States spends $50 billion per year due to poor maintenance of industrial facilities, and existing documentation is not enough to prevent deterioration of the industrial building stock (Thomas & Thomas 2018). This is a consequence of the aging industrial building stock. Buildings require periodic maintenance, especially those built in the 20th century, where the replacement of major systems is needed to avoid failure. To give an example, 72% of existing facilities in the United States are over

20 years old, requiring a high financial maintenance cost (The American Institute of Architects and Rocky Mountain Institute 2013). The life cycle of industrial facilities represents a high value in maintenance worldwide; the United States Pipeline and Hazardous materials Safety Administration has on record more than 10,000 failures in oil and gas pipelines in the country, which resulted in $6 billion financial losses (US Department of Transportation 2016). Poor maintenance of existing and growing factories is affecting the territory's economy. It is essential to highlight that U.S Oil plants are running at a capacity of 82%, reflecting a 10% decrease in the average production capacity. The extended problem results from the cutbacks of maintenance, which makes it harder to improve production for the future by (Sanicola, Seba 2021). In addition to that, there are time constraints on how long equipment and industrial elements can be stopped without affecting production. Production boundaries cannot be crossed without paying additional expenses. Given the market needs, this paper contributes to the use of advanced point cloud processing methods to alleviate the harm and minimize time spent on the creation of 3D representations of industrial parts, which can then accelerate maintenance and lower costs.

**3D modeling and digital twinning**. 3D digital models have gradually shifted the construction industry standards a step ahead from 2D drawings to highly detailed 3D object-based information systems (Migilinskas et al. 2013). Building Information Models (BIM) have been extensively used in the past decades for the construction of new buildings. However, the process of generating BIM for operating existing facilities remains cumbersome.

The first 3D BIM software was released in 1970. As a result, constructed buildings built before the 1970's lack the existence of 3D models that describe the design, functional, and operating systems of buildings and their components (Eastman et al. 1974). BIM has been implemented in the United Kingdom (UK) and the United States at a faster rate compared to other countries. For example, 73% of construction professionals in the UK reported using BIM (Statista 2022). However, there is a need to improve productivity when generating BIM models. The most widely used software modeling packages are Autodesk, Clearedge3D, and AVEVA, because they have tools that allow the 3D modeling of mechanical and piping equipment. Still, they require highly qualified labor to achieve accurate modeling results, since humans rely on manuallayeringandrecognitionofshapes. Thegeneration of a 3D model is a step forward towards the generation of a digital copy of the existing conditions of industrial facilities. The possible existing gap of using 3D models is the continuous representation of the life cycle of facilities.

In contrast to BIM, a Digital Twin (DT) is not just a digital copy of the infrastructure. It is a geometrically accurate replica of the existing conditions and it is able to show changes, while being connected to its physical counterparts (Sacks et al. 2020, IET 2019). The concept of DT was originally created by NASA, and it is defined as the continuous improvement in the generation of product designandengineeringpractices(Glaessgen&Stargel2012). A DT represents the life cycle of a facility, including detailed information that can promote proactive maintenance of the facility, lowering the cost of unknown failures to come. DTs incorporate a plethora of data sources, in order to represent existing facilities accurately. As an example, DTs allow builders to save 15-25% in the infrastructure market by 2025, because they help monitor the inefficiencies enabling accurate maintenance repairs (Barbosa et al. 2017, Gerbert et al. 2016).

3D data capturing of the geometry of existing industrial facilities plays a pivotal role in cost, and it is one of the most labor-intensive processes in the generation of a DT (Hullo et al. 2015). This process is known as geometric digital twinning (Agapaki & Brilakis 2021).

Speeding the process of 3D geometric digital twinning will result in immediate access to documentation, facilitating owners and engineering teams to gather information, improving the decision-making processes that can lead to proactive maintenance of industrial facilities. Also, implementing state-of-the-art deep learning algorithms can help reduce the 3D modeling time and cost. However, databases of labeled, virtual laser-scanned industrial facilities need to be developed to generate the input. That is why this paper proposes the generation of synthetic, labeled datasets to test whether this input can be used in existing point cloud processing algorithms by (Agapaki & Brilakis 2020). The milestone is to reduce labor-extensive work for easier adoption of deep learning algorithms. Class segmentation is an approach that can classify a remotely sensed image based on image segments (Agapaki & Brilakis 2020). The implementation of class segmentation contributes to the interpretation of objects with better accuracy regarding its components and condition of mechanical parts. An example of data that is successful by the implementation of class segmentation is the CLOI dataset. The point cloud dataset named CLOI is defined as a "novel class segmentation solution that consists of an optimized PointNET++ based deep learning network and post-processing algorithms that enforce stronger contextual relationships per point". The CLOI dataset is based on class segmentation generated by Terrestrial Laser Scanners, and it was developed to efficiently minimize the cost and manual labor during the generation of industrial objects (Agapaki & Brilakis 2020).

The current constraint is that real-world data is limited. The focus of this investigation is to generate synthetic data to test the ability of class segmentation algorithms on classifying laser scanned data obtained by virtual laser scanners.

## Background

Simulations are vital for finding suitable parameters of virtual laser scanners due to the lack of labeled point cloud data. However,thereisnocommercialvirtuallaserscanner that generates point clouds that include information on the existingconditionsofindustrialfacilities. Simulationtools developed in the literature for engineering and science purposes provide a tool for analyzing existing options of generating data by (Oden et al. 2006). They give space to improve the efficiency of point cloud segmentation by creating different scenarios (i.e., changing the parameters of a 3D model). Point cloud simulation requires two main inputs: (a) a 3D model of the facility to be scanned and (b) information regarding the sensor used.

The input 3D model should provide spatial relationships required to create a representative 3D point cloud. Simple triangulated mesh models are insufficient for automating the labeling of simulated point cloud data, because they lack the semantic information and component labels required to determine which class/object category each point should belong to. Synthetic point clouds or meshes generated from 3D models without the parameters of a real laser scanner are "perfect", but they cannot represent the actual real conditions. As a result, it is necessary to generate synthetic point clouds with parameters of real laser scanners. BIM models contain the 3D spatial and semantic data needed for automated labeling of simulated point cloud data. The Industry Foundation Classes (IFC) BIM format is a convenient, free, and interoperable to use format for this purpose by (Agapaki et al. 2018).

IFC is a data model that presents a metadata set established by buildingSMART. The IFC format includes

BIM representations with well-defined characteristics of building components. The IFC standard incorporates terms, concepts, and data specification elements. IFC is a format in which each element represents its geometric shape. The IFC format allows the modification of parameters during 3D modeling, giving the advantage of being implemented in research instead of other BIM representations by (Agapaki et al. 2018). 3D representations such as CAD 3D meshes could be "sparse and irregular" to manage. In this case, it is necessary to compare results to initial parameters to recover values to assess the accuracy of the method by (Bénière et al. 2013). It can be concluded that the IFC format allows users to have a smoother manipulation of 3D models.

3D representations of laser scanned data rely on understanding the simulation and computational needs for specific problems. Different simulations depend on the type of input scene model and how the interaction between the laser scanner and the object is modeled (Winiwarter et al. 2021). In the next sections, we analyze simulation platforms that can generate virtual laser scanned data with their parameters as well as existing synthetically generated point cloud datasets.

**Analysis of Virtual Laser Scanners**

The study focuses on finding the most precise virtual laser scanners with parameters that allow the modification of sensor specifications and settings to produce labeled point cloud data of industrial facilities. We focus on four laser scanners, of which three are virtual laser scanners named: Helios, Helios++, BlenSor; and one is a commercial simulation tool named: VirScan3D (Chizhova et al. 2021, Bechtold & Höfle 2016). VirScan3D is a terrestrial laser scanner simulator designed as an educational tool where independent users from university courses execute testing results. Yet, the results have to be evaluated to analyze its functionality. The goal of the laser scanner is to provide a tool for teaching purposes (Chizhova et al. 2021). The authors have not made the inner workings of the algorithms used in the VirScan3D software available. Therefore, it is excluded from further analysis.

The Helios library has a core package and various modules that perform tasks. The platform component simulates where the scanner is mounted with various parameters that can be adjusted to represent specific classes (Four-wheel ground vehicle with one steerable axle, helicopter/multicopter, simple linear interpolated movement along straight lines, and a dummy platform without movementcode). Thescenecomponentcontainsadatastructure that holds scene geometries and materials. A scene in Helios is defined as a triangle mesh that holds as reference a "material definition" able to describe physical properties of the scanned surface. Also, point cloud simulations use ray-casting for the transmission of laser pulses from the laser scanner to the scenes surfaces allowing a set of recommendations for optimal scanner positions. Finally, the scanner component is the actual scanner. On the other hand, Helios presents limitations on detecting collision, but it can capture movement such as the ones from vegetation with certain freedom. To date, the current state of the software only allows indirect control of the scanner by the XML configuration files, and the survey playback module. Future applications propose a graphical front-end to allow direct interaction with laser scanners. (Bechtold & Höfle 2016).

AnapplicationofHELIOSwasinvestigatedonasample industrial dataset by (Noichl et al. 2021). The study compared real laser scans and synthetically created point clouds. The differences presented in the results of both scenarios were due to the inaccuracies of the laser scanner itself. The primary concern is the deviation of the model and the appropriate format of geometric information representation. Data was less noisy than the real laser scanned point cloud, as a result less realistic. The study concluded that synthetic data used relies extensively on higher precision of the laser scanner because it will perform worse in real data.

As the evolution of Helios, a newer version of a laser scanner plays a pivotal role, called Helios++. Helios++, comparedtoHelios, presentsan83%reductioninruntimes and 94% less memory requirements. It simulates beam divergence using a subsampling strategy, and it creates fullwaveform outputs as the basis for analysis. HELIOS++ ensures that the angular distance between adjacent subrays is approximately constant, allowing the combination of input data from multiple sources and data formats. It is classified into two main categories: transmissive and nontransmissive. The transmissive pulse stops when an object with a dash is approaching, and the non-transmissive goes on until unlimited distance. This version allows a faster and easier generation of labeled training data. Data does not suffer from ground truth errors because when the network has learned to represent data, it can be modified to real data by adding small amounts of training data. As a result, it permits the generation of labeled point cloud data, which cuts the costs and provides the possibility of generating a massive amount of training data. However, files might not be supported by the current memory. In this case, a two-stage algorithm is implemented to process point clouds of "arbitrary size". The inputs quality is predominant for obtaining representative results because being physically realistic allows the estimation of accurate occlusion and point density. This versions process has a live preview of the point cloud acquisition in Python, giving a visual impression during simulations. However, the current drawback of Helios++ is the amount of energy needed for a single laser shot, affecting how realistic the point clouds are in terms of physical accuracy by (Winiwarter et al.

2021). A comparison between Helios and Helios++ capabilities is presented in Figure 1.

The third laser is BlenSor, which does not allow realtime simulation. It casts several rays simultaneously, and simulation accuracy increases by modifying the sensor code if features are not yet available for the ray-tracing required parameters. The speed of light is finite, allowing reflection and precision on most surfaces. Although functionality is limited because of the computational demand to simulate a large amount of laser rays, the performance suffers on the way. As a result of the high computational demand, the laser focuses on the simulation of the sensors instead of the sensor's interaction with the environment. One of the differences between BlenSor and Helios/Helios++ is the focus on data creation. Blensor is designed to create data offline to focus on usability and features. Blensor simplifies simulations related to unstable and dynamic scenarios such as car crashes. BlenSor can be used with a software called Blender because its engine allows the interaction of objects to complex simulated scenarios in applications with Blensor by (Gschwandtner et al. 2011).

## Point Cloud Class segmentation

Generally for image-based classification and segmentation, Convolutional Neural Networks (CNNs) are used due to their higher accuracy. They are extensively used in image segmentation by (Krizhevsky et al. 2012), text classification by (LeCun et al. 2008), and self-driving vehicles by (Teichmann et al. 2018). A CNN consists of convolutional and pooling layers to merge into local information per pixel. There are three main groups to classify the 3D deep learning methods. These are view-based (Su et al. 2015, Kalogerakis et al. 2017, Wei et al. 2016), volumetric (Klokov & Lempitsky 2017), and geometric deep learning methods (Qi et al. 2017).

Traditional CNNs are not useful to classify and adjust tovariationdensity,sincetheyonlyprocessstructureddata. To solve such a complex issue, our previous research used PointNET++ SFR (part of CLOI-NET) which is one of the best techniques in geometric deep learning methods. It

| | HELIOS (JAVA) Version 2018-09-24 | HELIOS ++ Version 1.0.0 | | | Visualization |
|---|---|---|---|---|---|
| Echo width | ✓ | ✓ | | | |
| Waveform output | ✓ | ✓ | ✓ | | |
| File Format | XYZ | XYZ | XYZ | LAS | |
| Scene 1 (ALS, GeoTIFF) | 4,712.1 ± 613.8 s  6,570 MB | 2,773.0 ± 53.8 s  2,246 MB | 2,403.8± 18.0 s  2,246 MB | 1,912.10 ± 20.22 s  2,246 MB | 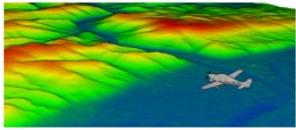 |
| Scene 2 (MILS, geometric primitives) | 68.0 ± 2.3 s  560 MB | 23.9 ±0.2 s  24 MB | 23.0±0.3 s  24 MB | 16.1±0.6 s  24 MB | 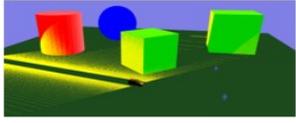 |
| Scene 3 (TLS, tree models) | 362.6± 6.7 s  5,360 MB | 97.2 ±1.3 s  314 MB | 74.4 ±1.0 s  314 MB | 62.4 ±0.3 s  314 MB | 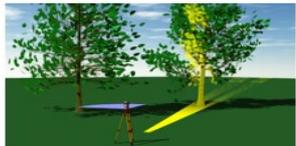 |

*Figure 1: Performance comparison of Helios and Helios ++ (default parameters for waveform modeling beamSampleQuality=3 and binSize$_{ns}$=0.25; numBins=100 and numFullwaveBins=200 for HELIOS++ and HELIOS, respectively). Run times are average of three runs (± standard deviation), memory footprint is the highest value (maximum) during the full run (Winiwarter et al, 2021).*

**Available Synthetic Point Cloud Datasets**

There is a wealth of synthetic point cloud datasets capturing various types of objects. These are captured with different types of virtual scanners such as Blensor by (Gschwandtner et al. 2011) and LIDAR and are summarized in Figure 2. A LIDAR sensor has flexible configurations to scan various types of objects. The parameters of the LIDAR sensor by (Wang et al. 2019) shown in Figure 2 are µ, ν, α, β. The α and µ are in the X-Z dimension representing vertical scanning. ν and β are in the X-Y dimension representing horizontal scanning.

extracts local features capturing both local and global fine geometric structures from small neighborhoods. These local features are further grouped into larger units and combine their outputs by using a hierarchical approach to produce higher level features. This process is repeated until we obtain the features of the whole point set.

| Virtual Scanner | Number of classes | Visualization |
|---|---|---|
| **Blensor**<br>**Scan type:** Generic lidar<br>**Max distance:** 100 m<br>**Angle resolution:** 0.05 m<br>**Start angle:** −180°<br>**End angle:** 180°<br>**Frame time:** 1/24 s | 9 (Building, Car, Natural Ground, Ground, Pole Like, Road, Street Furniture, Tree, Pavement) | 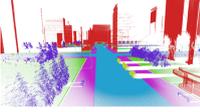<br>*class labels*<br>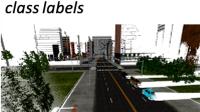<br>*RGB values*<br>Griffiths, D. and Boehm, J. (2019) |
| **LiDAR**<br>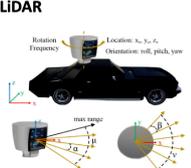 | 10 (Building, Road, Vegetation, Sidewalk, Vehicle, Pedestrian, Pole, Wall, Fence, Traffic Sign) | 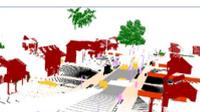<br>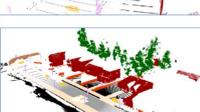<br>Wang et al. (2019) |

*Figure 2: Literature review on available synthetic datasets*

The objective of this paper is to derive whether syntheticpointclouddataisbeneficialforpointcloudsegmentation. This will be achieved by a Helios, Helios ++, and Blensor qualitative comparison. Still, Helios++ presents a higher virtual laser scanning performance. It provides flexible computational requirements and a better approximation to physical realism.It is not possible to compare synthetic data parameters that are realistic over laser scanners. That is why Helios++ is selected to resemble the conditions. Also, Helios++ is able to simulate laser scanned datainrealtime. Itsobjectmodelcalledtransmissivevoxel is an achievement to virtual laser scanning characteristics because of the ability to penetrate objects in the absence of detailed 3D mesh models. Another positive development is the maximum detection binning mechanism, even if waveforms are not written to a specific output file. Also, Helios++ allows the usage of significant input of point clouds by dividing it into smaller parts. As a result, the generation of VLS (virtual laser scanning) realistic data invites users to rely on the algorithm. Therefore, it is most suitable for our approach.

## Research Approach

Figure 3 presents our research methodology. Helios++ can be easily used via the command line or Python bindings. It can be installed and setup in numerous ways. It is recommended that the user set up a virtual environment to download and install all necessary packages and then start using the software as desired. The most optimum way of installing Helios++ would be with Docker. In our research, we make use of python bindings to run experiments and record the results.

Once the software is running, we then use the object file on Helios++. It is essential that the input is in .obj format due to the lack of support with other formats. Our first step is to convert the object file into an XML file. The XML file acts as a survey file which consists of Platforms, Scanners and Scenes. These parameters are determined by the type of object model provided. This laser scan XML file can be a Terrestrial (TLS), Airborne (ALS), Mobile (MLS) and UAV-borne (ULS). The scene is made up of components that are scanned, the platform which in turn is made up of groundplane.obj found in the repository and acts as a virtual scanner and lastly, the scanner itself. The survey consists of various legs which are wavepoints for the platform. An illustration of the elements is presented in Figure 4.

Noichl et al. (2021) uses Helios to create the point cloud model. The scene and survey preparation are similar for Helios and Helios++. Both require a .obj file with information and the orientation being stored in an XML file. Helios++ lacks a UI but makes up for it with better computation. Helios++ has a better accuracy over the cost of run time or vice-versa depending on the parameters set. For the output, both Helios and Helios++ create one point cloud per scan. All the outputs can be combined to formonesinglepointcloud. TocreatetheXMLfile, werun the python script for the scene generator with the following arguments: the folder that contains the obj files, path to the ground plane, name of the scene to be generated, number of objects to be distributed, number of circular segments and radius of the scan positions around center of the scene.

Once we generate the XML, we get the survey and scene parameters in one file. We also need to separate the scene and survey into 2 XML files as the script for generating the point cloud does not detect the scene. We add the tag "<?xml version=1.0?>" to both our survey and scene files. The filters applied onto the scene are rotation and translation. We remove the rotation filters so that our object lay flat on the ground plane. In future work, we intend to address that limitation. After creating the appropriate XML, we then proceed with generating the point cloud for the object/scene of interest. The point cloudfileconsistsofthefollowingparameters: X,Y,Z,Nx, Ny and Nz co-ordinates, Scalar values (representing class labels) as well as R, G and B values. The output consists of a point cloud extension(.xyz) file and the log files that are automatically stored in the outputs folder. If needed,

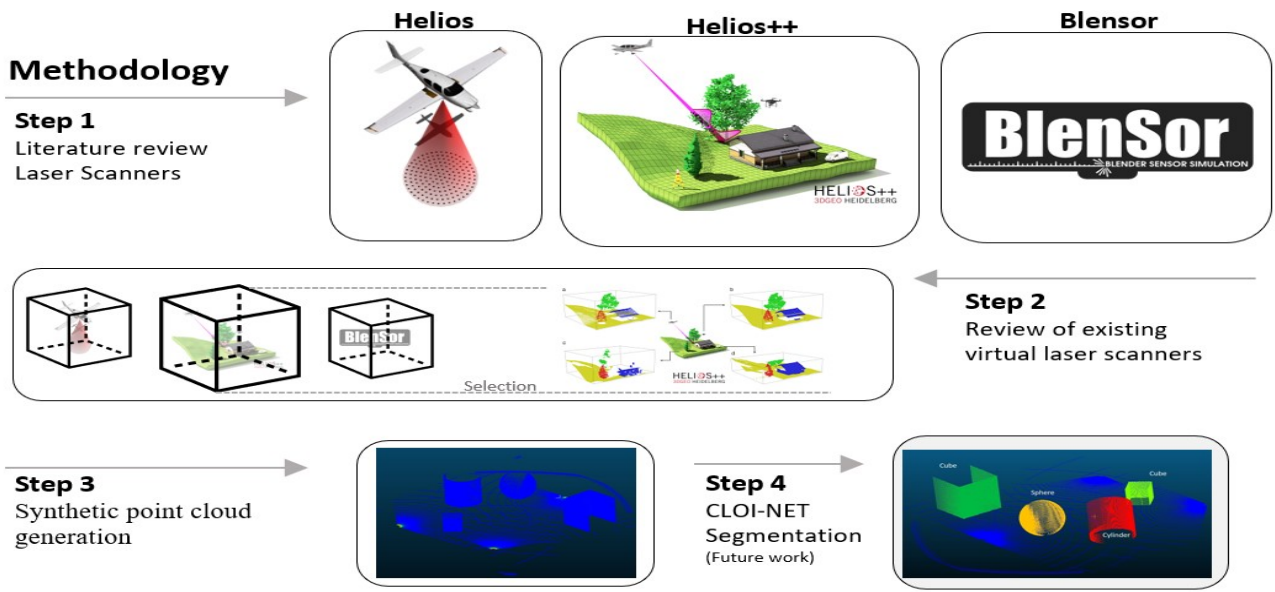

*Figure 3: Proposed Research Methodology*

one can visualize the waveform by providing specific flags
during run time. We run the script to generate the point cloud and then compare the object file with the point cloud model.

We generate another point cloud for the basic joints and edges. Here, we define the number of circular segments as 14, number of defined objects as 5 and we keep the radius 100, which in hindsight is large enough for our scene. Some sample scene results are illustrated in Figures 5-7.

**CLOI-NET Segmentation**

In this methodology by (Agapaki & Brilakis 2020), the first step is data pre-processing and converting it into the file format which will be suitable for running CLOI-NET which is explained below.

After completing the synthetic point cloud generation step, we get the point cloud data in (.xyz format). We then convert it into (.txt) format, so that data pre-processing can be performed. We only use X, Y, Z labels in the CLOINET model. We use data pre-processing technique and convert it into batches of small files using a 3D sliding window/block approach. Once the data is prepared, the second step in the CLOI-NET method is to predict a class labelperpointusingamodifiedversion(SFR-Smallerand Fewer neighborhoods with smaller Radius) to accurately segment the CLOI shapes. **Results**

---

**Synthetic Point Cloud Generation**

We previously defined the parameters required to generate the XML file. For the point cloud generation of the shapes, we define the number of circular segments as 3, number of defined objects as 4 and the radius of scanned positions around the center of the scene to be 50.

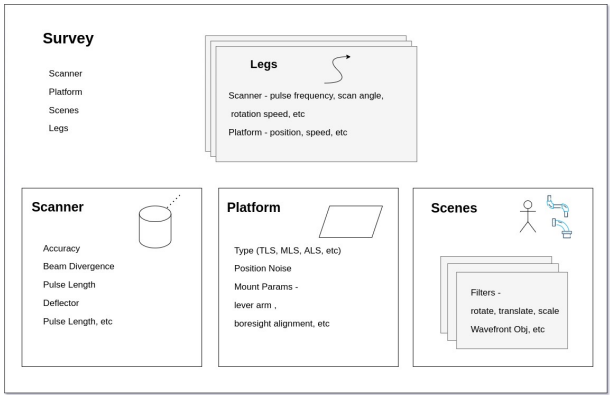

Figure 4: Elements of the HELIOS++ platform

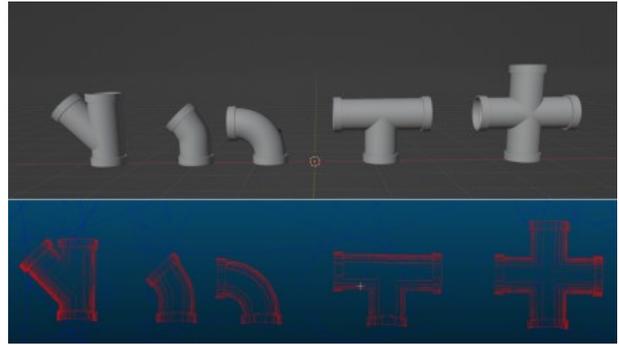

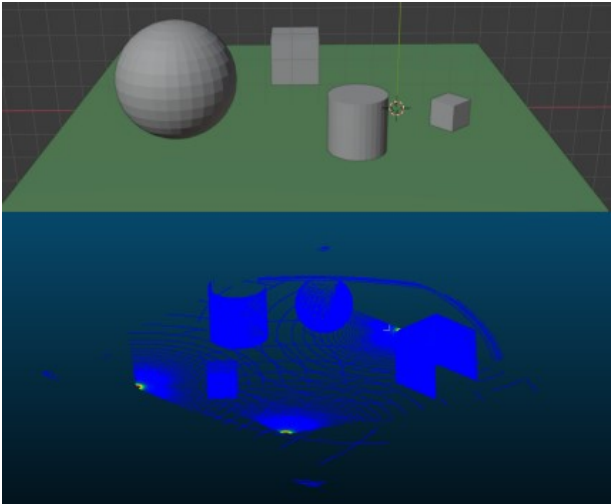

Figure 5: Sample scene (input) and synthetically generated laser scan (output)

## Conclusions

Our analysis illustrated that HELIOS++ has potential to be used for point cloud class segmentation, however further

Figure 6: Sample scene with elbow models (input) and synthetically generated laser scan (output)

performance.

research is needed to simplify the point cloud generation and using the generated outputs for training the existing segmentation networks as illustrated above. Another limitation is the significant manual user input needed to segment the IFC files that need to be parsed to individual objects, so that they can be used in the HELIOS++ virtual scanner platform. HELIOS++ minimizes the needs for data conversion that lead to a more efficient generation of virtual scanned point clouds. The significance of this work is to minimize human labor involved in the generation of annotated point cloud datasets without sacrificing

## Acknowledgments

We thank the Florida Space Institute (NASA) for sponsoring this research. We gratefully acknowledge the collaboration of all academic and industrial project partners. Any opinions, findings and conclusions or recommendations
expressed in this material are those of the authors and do not necessarily reflect the views of the institutes mentioned

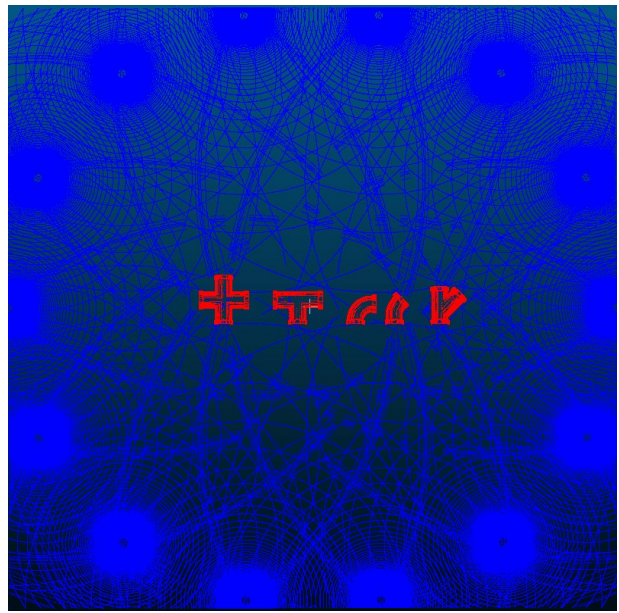

*Figure 7: Top view of the sample scene with expected class segmentation labels*

above.

# References

Agapaki, E. & Brilakis, I. (2020), 'Cloi-net: Class segmentation of industrial facilities point cloud datasets', *Advanced Engineering Informatics* **45**, 101121.

Agapaki, E. & Brilakis, I. (2021), 'Cloi: An automated benchmark framework for generating geometric digital twins of industrial facilities', *arXiv preprint arXiv:2101.01355*.

Agapaki, E., Miatt, G. & Brilakis, I. (2018), 'Prioritizing object types for modelling existing industrial facilities', *Automation in Construction* **96**, 211–223.

Barbosa, F., Woetzel, J. & Mischke, J. (2017), "reinventing construction: A route of higher productivity", Technical report, McKinsey Global Institute.

Bechtold, S. & Höfle, B. (2016), 'Helios: A multi-purpose lidar simulation framework for research, planning and training of laser scanning operations with airborne, ground-based mobile and stationary platforms', *ISPRS Annals of Photogrammetry, Remote Sensing & Spatial Information Sciences* **3**(3).

Bénière, R., Subsol, G., Gesquière, G., Le Breton, F. & Puech, W. (2013), 'A comprehensive process of reverse engineering from 3d meshes to cad models', *Computer Aided Design* **45**(11), 1382–1393.

Chizhova, M., Gorkovchuk, D., Kachkovskaya, T., Popovas, D., Gorkovchuk, J., Luhmann, T. & Hess, M. (2021), 'Qualitative testing of an advanced terrestrial laser scanner simulator: users experience and feedback', *The International Archives of Photogrammetry, Remote Sensing and Spatial Information Sciences* **43**, 29–35.

Eastman, C. et al. (1974), 'An outline of the building description system. research report no. 50.'.

Gerbert, P., Castagnino, S., Rothballer, C., Renz, A. & Filitz, R. (2016), 'Digital in engineering and construction', *The Boston Consulting Group* pp. 1–22.

Glaessgen, E. & Stargel, D. (2012), The digital twin paradigm for future nasa and us air force vehicles, in '53rd AIAA/ASME/ASCE/AHS/ASC structures, structural dynamics and materials conference 20th AIAA/ASME/AHS adaptive structures conference 14th AIAA', p. 1818.

Gschwandtner, M., Kwitt, R., Uhl, A. & Pree, W. (2011), "blensor: Blender sensor simulation toolbox", in 'International Symposium on Visual Computing', Springer, pp. 199–208.

House, W. (2021), 'Fact sheet: The bipartisan infrastructure deal'.

Hullo, J.-F., Thibault, G., Boucheny, C., Dory, F. & Mas, A. (2015), 'Multi-Sensor As-Built Models of Complex Industrial Architectures', *Remote Sensing* **7**(12), 16339–16362.

IET (2019), Digital Twins for the built environment, Technical report.

Institute, M. G. (2021), 'Building a more competitive us manufacturing sector'.

Kalogerakis, E., Averkiou, M., Maji, S. & Chaudhuri, S. (2017), 3d shape segmentation with projective convolutional networks, in 'proceedings of the IEEE conference on computer vision and pattern recognition', pp. 3779–3788.

Klokov, R. & Lempitsky, V. (2017), "escape from cells: Deep kd-networks for the recognition of 3d point cloud models", in 'Proceedings of the IEEE International Conference on Computer Vision', pp. 863–872.

Krizhevsky, A., Sutskever, I. & Hinton, G. E. (2012), 'Imagenet classification with deep convolutional neural networks', *Advances in neural information processing systems* **25**, 1097–1105.

LeCun, Y., Boser, B., Denker, J. S., Henderson, D., Howard, R. E., Hubbard, W. & Jackel, L. D. (2008), 'Backpropagation applied to handwritten zip code recognition', *Neural computation* **1**(4), 541–551.

McKinsey (2016), 'Bridging global infrastructure gaps'.

Migilinskas, D., Popov, V., Juocevicius, V. & Ustinovichius, L. (2013), 'The benefits, obstacles and problems of practical bim implementation', *Procedia Engineering* **57**, 767–774.

Miyamoto, K. & Wu, Y. (2018), 'Enhancing connectivity through transport infrastructure: The role of official development finance and private investment'.

Noichl, F., Braun, A. & Borrmann, A. (2021), 'bim-toscan' for scan-to-bim: Generating realistic synthetic ground truth point clouds based on industrial 3d models, in 'Proceedings of the 2021 European Conference on Computing in Construction'.

Oden, J., Belytschko, T., Fish, J., Hughes, T., Johnson, C., Keyes, D., Laub, A., Petzold, L., Srolovitz, D., Yip, S. et al. (2006), 'Simulation-based engineering science: Revolutionizing engineering science through simulation', *NSF Blue Ribbon Panel on SBES* .

Organization for Economic Co-operation and Development (2018), 'China's belt and road initiative in the global trade, investment and finance landscape'.

Qi, C. R., Su, H., Mo, K. & Guibas, L. J. (2017), Pointnet: Deep learning on point sets for 3d classification and segmentation, *in* 'Proceedings of the IEEE conference on computer vision and pattern recognition', pp. 652–660.

Sacks, R., Brilakis, I., Pikas, E., Xie, H. S. & Girolami, M. (2020), 'Construction with digital twin information systems', *Data-Centric Engineering* **1**.

Sanicola, Seba (2021), 'Lack of overhauls at u.s. refiners could stall industry recovery'.

Statista(2022), 'Bimadoptionrateinconstructionindustry in the united kingdom 2011-2020'.

Su, H., Maji, S., Kalogerakis, E. & Learned-Miller, E. (2015), Multi-view convolutional neural networks for 3d shape recognition, *in* 'Proceedings of the IEEE international conference on computer vision', pp. 945–953.

Teichmann, M., Weber, M., Zoellner, M., Cipolla, R. & Urtasun, R. (2018), 'Multinet: Real-time joint semantic reasoning for autonomous driving', pp. 1013–1020.

The American Institute of Architects and Rocky Mountain Institute (2013), 'Deep energy retrofits: 1981 an emerging opportunity.'.

Thomas, D. S. & Thomas, D. S. (2018), *"The costs and benefits of advanced maintenance in manufacturing"*, US Department of Commerce, National Institute of Standards and Technology.

US Department of Transportation (2016), 'Distribution, transmission & gathering, lng, and liquid accident and incident data. pipeline and hazardous materials safety administration'.

Wang, F., Zhuang, Y., Gu, H. & Hu, H. (2019), 'Automatic generation of synthetic lidar point clouds for 3-d data analysis', *IEEE Transactions on Instrumentation and Measurement* **68**(7), 2671–2673.

Wei, L., Huang, Q., Ceylan, D., Vouga, E. & Li, H. (2016), Dense human body correspondences using convolutional networks, *in* 'Proceedings of the IEEE conference on computer vision and pattern recognition', pp. 1544–1553.

Winiwarter, L., Pena, A. M. E., Weiser, H., Anders, K., Sánchez, J. M., Searle, M. & Höfle, B. (2021), 'Virtual laser scanning with helios++: A novel take on ray tracing-based simulation of topographic full-waveform 3d laser scanning', *Remote Sensing of Environment* p. 112772.